\documentclass[runningheads]{llncs}
\usepackage{graphicx}
\usepackage[utf8]{inputenc} 
\usepackage{amsmath} 
\usepackage{amssymb}
\usepackage{bm}
\usepackage[normalem]{ulem}
\usepackage{txfonts}
\usepackage{graphicx,subfig}
\usepackage{float}
\usepackage{caption}
\usepackage{hyperref}
\usepackage{CJKutf8}
\usepackage{tikz} 
\usepackage{hyperref}
\usepackage{sidecap} 
\usepackage{microtype}
\usepackage{booktabs}
\usepackage{upquote}
\usepackage{listings} 
\usepackage{tikz}
\usetikzlibrary{arrows.meta}
\usetikzlibrary{bayesnet} 
\usepackage[boldmath]{numprint} 
\usepackage{pifont} 
\usepackage{array}
\usepackage{makecell} 
\usepackage{mathtools}
\usepackage{calrsfs}
\DeclareMathAlphabet{\pazocal}{OMS}{zplm}{m}{n}
\newcommand{\defeq}{\vcentcolon=}
\newcolumntype{H}{>{\setbox0=\hbox\bgroup}c<{\egroup}@{}}

\definecolor{darkgray}{gray}{0.60}

\begin{document}
\title{Errors-in-variables Modeling of Personalized Treatment-Response Trajectories}
%
%
\author{Guangyi Zhang\inst{1} \and
Reza Ashrafi\inst{1} \and
Anne Juuti\inst{2} \and
Kirsi Pietiläinen\inst{2} \and
Pekka Marttinen\inst{1}}
\authorrunning{G. Zhang et al.}
%
\institute{Aalto University, Finland\\
\email{\{guangyi.zhang,reza.ashrafi,pekka.marttinen\}@aalto.fi} \and
University of Helsinki, Finland\\
\email{anne.juuti@hus.fi,kirsi.pietilainen@helsinki.fi}}
\maketitle              
\begin{abstract}
Estimating the effect of a treatment on a given outcome, conditioned on a vector of covariates, is central in many applications. However, learning the impact of a treatment on a continuous temporal response, when the covariates suffer extensively from measurement error and even the timing of the treatments is uncertain, has not been addressed. We introduce a novel data-driven method that can estimate treatment-response trajectories in this challenging scenario. We model personalized treatment-response curves as a combination of parametric response functions, hierarchically sharing information across individuals, and a sparse Gaussian process for the baseline trend. Importantly, our model considers measurement error not only in treatment covariates, but also in treatment times, a problem which arises in practice for example when treatment information is based on self-reporting. In a challenging and timely problem of estimating the impact of diet on continuous blood glucose measurements, our model leads to significant improvements in estimation accuracy and prediction.

\end{abstract}
%
%
%
\section{Introduction}
Increasing popularity of electronic health records (EHRs) and smart healthcare services has led to accumulation of large quantities of heterogeneous data, with potential to considerably improve the efficiency of clinical practice and health services \cite{Powell2005}. 
This highlights the importance of novel machine learning techniques for EHR data, which can be integrated with mobile apps to provide personalized guidance for purposes ranging from early diagnosis to support for lifestyle change \cite{OpporEHR,DeepEHR}. The latter is specifically relevant to reduce the cost of chronic diseases in the face of the aging population. For instance, the annual economic cost of diabetes in the U.S. is approximately \$250 billion \cite{ADA2013}. 

Inferring relationships between correlated variables is essential in many fields. An important question is to estimate a patient’s response to a given treatment, comparing the patient's data from before and after the treatment. A traditional solution is to use randomized controlled trials, which, however may be infeasible due to the cost or ethical considerations. One possibility is to use mechanistic models, specifically tailored for the problem, and use data to learn about their unknown parameters. However, these models require substantial expert knowledge and are not applicable if the underlying mechanism is unknown. On the other hand, data-driven methods, trained on observational EHR data, provide a promising alternative.

Estimating the impact of a treatment is particularly challenging when the response is a continuous curve consisting, for example, of a time-series of measurements of a biological marker. In such cases, the outcome is typically modeled by a Gaussian process \cite{rasmussen2004gaussian}, but also neural networks have been considered \cite{lim2018forecasting}. The treatment may either be a continuous dose function \cite{soleimani2017treatment}, or a discrete event in time \cite{Yanbo16,schulam2017reliable}. The latter approach is often relevant in practice when treatments are recorded as discrete events, even if their true duration is not exactly zero. Treatment data are usually sparse, and hence it is essential to share relevant information in a probabilistic model. A latent trajectory model of \cite{schulam2015framework} uses additive components to explain variation on population and individual levels. Conditional random fields can be incorporated to further capture correlations between different treatment types \cite{schulam2016integrative}, and multivariate response curves can be modeled by learning latent structure \cite{soleimani2017treatment} shared across the outcomes. 

Despite increased recent attention, there are still crucial issues in treatment-response estimation that have not been addressed when the response is continuous. Most importantly, the treatments are consistently assumed to be exactly measured and known, while in reality the treatment input may be severely perturbed by numerous factors. This problem dramatically escalates for user-reported treatment data, which potentially results in complete discredit of the findings \cite{Kreider2010}. The erroneous effect is two-fold, i.e., in addition to measurement error in covariates, the actual time of the treatment might be known only approximately. Another issue arises from the competing relationship between a counterfactual trend, i.e., the evolution of the outcome assuming no treatment, and the treatment response. When modeled and trained jointly, it easily happens that a flexible trend completely overrides the treatment response, and therefore these two components are often trained separately in practice.

To address the mentioned shortcomings, we introduce errors-in-variables (EIV) framework for modeling of continuous treatment-response trajectories. The EIV models account for measurement errors not only in the output variable, as common regression, but also in the inputs \cite{Griliches74,GRILICHES198693}. They are closely related to latent-variable models in machine learning \cite{murphy2013Machine,bishop2006pattern}, and based on modeling the unobserved true values from which noisy observations are obtained. Our contributions can be summarized as follows:
\begin{itemize}

  \item We formulate an EIV model for personalized treatment-response trajectories, where a treatment comprises a vector of noisy covariates and treatment times are uncertain.
  
  \item We introduce an interpretable hierarchical prior on the treatment effects that efficiently shares information between individuals, and allows training the full model jointly, appropriately balancing between the trend and the responses.
  
  \item In a challenging topic, representative of the current technological mega-trend on self-monitoring data from wearable devices, we show our method can meaningfully estimate the personalized impact of diet on continuous blood glucose measurements.
  
\end{itemize} 

The code and data used in the analyses are available at [link added upon acceptance, reviewers can view the material in Supplement] and allow fully reproducing our results.

\section{Related work} 
\textbf{Treatment response}: Besides machine learning, the problem of treatment response estimation has been studied in various fields, including informatics for medicine and social sciences, where the data-driven approach can bring advantages compared to the experimental trials \cite{Passos2018}. For example, individual-level treatment response prediction has been studied for schizophrenia \cite{Cao2018} and depression \cite{Pearson2018}. An empirical comparison of classifiers for treatment-response prediction for chemoradiotherapy appears in \cite{Deist2018}. Topics studied in social sciences include the effect of a discrete treatment, years of education, on an individuals' income \cite{Card99}, and allowing a response to depend on social interactions and treatments for other individuals \cite{Manski2013}.

\textbf{Mechanistic models}: In contrast to the data-driven approaches used in machine learning, mechanistic models use substantial knowledge of a specific problem to characterize the system with differential equations, and inference is done for example using filtering algorithms. Similar to our application, \cite{Balakrishnan2014} and \cite{albers2017personalized} study blood glucose dynamics, affected by nutrition and other factors. Another example is a computational model of the physiological mechanisms for type-2 diabetes, aiming to quantify factors useful for prevention of the disease \cite{Joydeep2018}.

\textbf{EIV models for treatment response}: In contrast to the vast body of research concerning data-driven methods for the prediction of a treatment response, very little is known about their performance when measurement error is present. Authors of \cite{YulaiZhang14} study a method for inferring causal directions using EIV models, but they do not focus on the response conditioned on specific treatments. Regression on various student covariates, incorporating EIV, has been used to predict standardized test scores \cite{Lockwood14}. In \cite{IZA2010}, the effect of measurement error on a binary treatment response is analyzed, underlining the devastating impact of ignoring such errors. A model for measurement errors has been used to quantify uncertainty in order to increase the confidence in detecting genuine treatment changes for liver metastases \cite{Pathak2017}.

None of these works address the problem of estimating the impact of a multivariate vector of covariates on a continuous response with measurement error in covariates and uncertainty in treatment timing, the topic of this paper.

\section{Methods} 
In this section, we first review EIV models on a general level. Then we describe three essential components of our model for personalized treatment-response trajectories: a hierarchical prior on parametric treatment-responses functions, a Gaussian process model for the trend, and measurement error models. Throughout the section, we present the model in generic terms, but also outline the specific model that we use in Section \ref{ss:realdata} to estimate the impact of diet, recorded as nutrient contents of different meals, on continuous blood glucose measurements.

Our model is fully Bayesian, yielding uncertainty estimates for all parameters, essential in scientific applications. Inference is done using Markov chain Monte Carlo (MCMC) with the state-of-the-art No-U-Turn (NUTS) sampler \cite{hoffman2014no} implemented in software PyMC3 \cite{salvatier2016probabilistic}. Implementation details are discussed below and in the supplementary, and can be viewed in full in the published code.

\subsection{Errors-in-variables models}\label{ss:eiv}
\begin{figure}[tb]
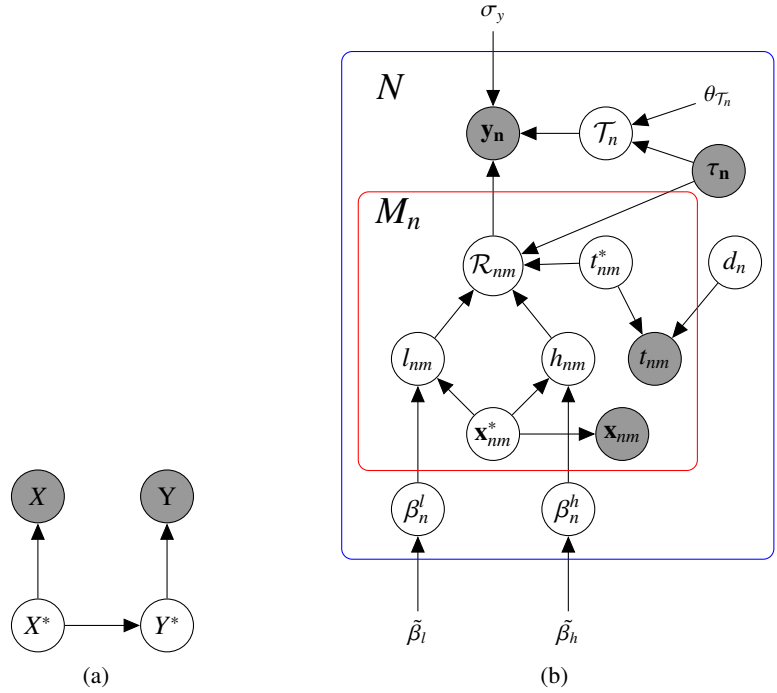

	\centering
	\subfloat[\mbox{}]
	{
		\makebox[.3\columnwidth]{
		  \centering 
  \tikz{ 
    \node[obs,fill={darkgray}] at(0,0) (X) {$X$} ;
    \node[latent,below=of X]  (Xs) {$X^*$} ;
    \node[latent, right=of Xs] (Ys) {$Y^*$} ;
    \node[obs, above=of Ys,fill={darkgray}] (Y) {Y} ;
    \edge {Xs} {X} ;
    \edge {Xs} {Ys} ;
    \edge {Ys} {Y} ;
  } 
	}\label{diag:eiv}} 
	\hspace{1mm}
	\subfloat[]{
		\makebox[.65\columnwidth]{ 
		\tikz{
    \node[obs,fill={darkgray}] at(0,0) (y) {$\mathbf{y_n}$} ;  
    \node[above=of y] (sigy) {$\sigma_y$} ;  
    \node[latent] at(1.5,0) (trend) {$\pazocal{T}_n$} ;  
    \node[obs,fill={darkgray}] at(3,-0.5) (t) {$\mathbf{\tau_n}$} ; 
    \node[] at(3,0.5) (theta_trend) {$\theta_{\pazocal{T}_n}$} ;  
    \node[latent, below=of y] (R) {$\pazocal{R}_{nm}$} ; 
    \node[latent, below=of trend] (tstar) {$t_{nm}^*$} ; 
    \node[latent, right=of tstar] (d) {$d_{n}$} ;  
    \node[latent] at(-1,-3) (l) {$l_{nm}$} ; 
    \node[latent] at(1,-3) (h) {$h_{nm}$} ; 
    \node[obs,fill={darkgray}] at(2.15,-3) (tt) {$t_{nm}$} ;  
    \node[latent] (Xstar) at(0,-4) {$\mathbf{x}_{nm}^*$} ;  
    \node[obs, right=of Xstar,fill={darkgray}] (X) {$\mathbf{x}_{nm}$} ;  
    \node[latent] at(-1,-5) (betaL) {$\beta_{n}^{l}$} ;  
    \node[latent] at(1,-5) (betaH) {$\beta_{n}^{h}$} ;  
    \node[below=of betaL] (betaLt) {$\tilde{\beta_{l}}$} ;  
    \node[below=of betaH] (betaHt) {$\tilde{\beta_{h}}$} ;  
    \edge {sigy} {y} ; 
    \edge {trend} {y} ; 
    \edge {t} {trend} ; 
    \edge {theta_trend} {trend} ; 
    \edge {t,tstar} {R} ;  
    \edge {R} {y} ; 
    \edge {d,tstar} {tt} ;  
    \edge {h,l} {R} ;  
    \edge {Xstar} {X} ;  
    \edge {Xstar} {h,l} ;  
    \edge {betaL} {l} ;  
    \edge {betaH} {h} ;  
    \edge {betaLt} {betaL} ; 
    \edge {betaHt} {betaH} ; 
    \tikzset{plate caption/.append style={above=2pt of #1.north west}} 
    \plate[color=blue, inner sep=0.3cm, xshift=-0.15cm, yshift=0.0cm] {plate1} {(y) (t) (R) (tstar) (d) (l) (h) (tt) (Xstar) (X) (betaL) (betaH)} {\Large{$N$}}; 
    \plate[color=red, inner sep=0.1cm, xshift=0.1cm, yshift=-0.2mm] {plate1} {(R) (tstar) (l) (h) (tt) (Xstar) (X)} {\hspace{2mm}\Large{$M_n$}}; 
  } 
	}\label{diag:model}} \\
\caption{a) General formulation of the EIV model. For clarity, parameters associated with the distributions are not shown. (b) Model for personalized treatment-response trajectories. Details of the model are discussed in the text. 
}
\end{figure} 

EIV models, a.k.a. measurement error models, are regression or classification models that, in contrast to most existing models, account for errors not only in the output variable but also in inputs \cite{carroll2006measurement,schennach2007nonparametric,MeasurementErrorBook,Fuller87}. Though commonly neglected, input mismeasurement may be extremely harmful. For example in simple linear regression it leads to biased estimates that can not be corrected for even with an infinite sample, while, on the other hand, unbiased homoscedastic error in the output variable only induces additional variability \cite{carroll2006measurement}. 
A graphical model for a general EIV model is presented in Figure \ref{diag:eiv}, where $X^*$ and $Y^*$ represent the true values of the inputs and the output, and $X$ and $Y$ are the corresponding noisy observations. The most important type of mismeasurement is \emph{classical error}, which corresponds to independence of an error term from the true value.

Except for the simplest case of linear regression \cite{pal1980consistent}, 
EIV modeling almost always requires auxiliary information or data to correct the mismeasurement bias in estimation. For problems that have an analytical solution, the bias can be corrected by a multiplication or addition of external terms \cite{hwang1986multiplicative}, e.g., an estimated reliability ratio \cite{carroll2006measurement}. For nonlinear models, auxiliary data, e.g., instrumental variables or repeated measurements, can be exploited to help correct bias, e.g., by estimating the density function of the true variable using a deconvolution technique \cite{schennach2007nonparametric}. However, without additional data, Bayesian EIV modeling is currently the most powerful and flexible approach, as it allows incorporating additional information in the form of distributional assumptions \cite{MeasurementErrorBook}. In this work, we adopt the Bayesian approach.

Mathematically, the measurement error mechanism is defined as the distribution of the noisy observed input, $X$, given the true unobserved input, $X^*$. The joint distribution of the model factorizes accordingly as:
\begin{equation}
    P(X^*,Y^*,X,Y,\Theta) = P(X|X^*,\theta_M) P(Y|Y^*,\theta_N) P(Y^*|X^*,\theta_R) P(X^*|\theta_E)P(\Theta),
\end{equation}
where $P(X|X^*,\theta_M)$ and $P(Y|Y^*,\theta_N)$ are called \textit{error} or \textit{measurement models}, $P(Y^*|X^*,\theta_R)$ is a \textit{response} or \textit{outcome model}, $P(X^*|\theta_E)$ is an \textit{exposure model}, and $\Theta=(\theta_M,\theta_N,\theta_R,\theta_E)$ are the corresponding parameters. Bayes theorem can be used to infer the unknown parameters and unobserved true values of the variables. 
\begin{equation}
    P(X^*,Y^*,\Theta|X,Y) \propto P(\Theta) \prod_i^N P(X_i|X_i^*,\theta_M) P(Y_i|Y_i^*,\theta_N) P(Y_i^*|X_i^*,\theta_R) P(X_i^*,\theta_E).
\end{equation}
If the exposure model is noninformative and the measurement model is symmetric, i.e., $P(X_i|X_i^*,\theta_M) = P(X_i^*|X_i,\theta_M)$, then the Bayesian modeling of classical error is equivalent to another class of mismeasurement techniques know as \emph{Berkson error modeling} \cite{carroll2006measurement}.
\begin{equation}
    P(X^*,Y^*,\Theta|X,Y) \propto P(\Theta) \prod_i^N P(X_i^*|X_i,\theta_M) P(Y_i|Y_i^*,\theta_N) P(Y_i^*|X_i^*,\theta_R). \nonumber
\end{equation}

A well-known difficulty with EIV models is that they are often nonidentifiable \cite{MeasurementErrorBook}, i.e. there are more than one set of values for the unknowns leading to the same model. This can be understood intuitively by noticing that the model stays the same if we multiply the linear regression coefficients by a constant factor and at the same time divide the estimated true values of inputs by the same factor. Therefore, to achieve identifiability, some crucial information about measurement model has to be assumed or estimated, e.g., the variance of a classical additive error in simple linear regression \cite{carroll2006measurement}.
The Bayesian paradigm offers a unique solution to the nonidentifiability of the EIV models, as long as mismeasurement is modest and the prior is sufficiently good \cite{gustafson2001case}. 

\subsection{Model for treatment-response trajectories}\label{subsect:treatment}

\textbf{Notation:} A graph of our model for treatment-response trajectories is presented in Figure \ref{diag:model}. We assume there are $N$ patients, and a trajectory consisting of a time series of length $G_n$ of the outcome (e.g. blood glucose) is observed for each individual:
\begin{equation*}
\mathbf{y_n} = (y_{n1},\ldots,y_{nG_n})^T, n=1,\ldots,N.
\end{equation*}
These measurements have been taken at times
\begin{equation*}
\mathbf{\tau_n} = (\tau_{n1},\ldots,\tau_{nG_n})^T, n=1,\ldots,N.
\end{equation*}
Furthermore, each patient has $M_n$ observed treatments (e.g. meals eaten), indexed by $m \in {1,\ldots,M_n}$, where each treatment is characterized by $P$ covariates:
\begin{equation*}
\mathbf{x}_{nm}=(x_{nm1},\ldots,x_{nmP})^T, \text{for all } m,n,
\end{equation*}
and the corresponding recorded treatment times are
\begin{equation*}
    \mathbf{t}_{n} = (t_{n1},\ldots,t_{nM_n})^T, \text{for all } n.
\end{equation*}
Here, $\textbf{x}_{nm}$ and $t_{nm}$ are assumed to be noisy observations of the treatment covariates and timings, and their true unobserved values are denoted by $\textbf{x}^*_{nm}$ and $t^{*}_{nm}$, respectively.

\textbf{Outcome model:} We model the observed outcome trajectory of individual $n$, $\mathbf{y}_n$, as
\begin{equation*} 
    \mathbf{y}_{n} = \pazocal{T}_n+\sum_m \pazocal{R}_{nm} + \textbf{e},
\end{equation*}
where $\pazocal{T}_n\in\mathbb{R}^{G_n}$ is a counterfactual trend (i.e. it describes the evolution of the outcome had the treatment not been taken), $\pazocal{R}_{nm}\in\mathbb{R}^{G_n}$ is the additive response to the $m$th treatment, and $\textbf{e}=(e_1,\ldots,e_{G_n})^T$ is the vector of errors with $e_i \sim N(0,\sigma_y^2)$. We note that the sum of the trend and the responses can be viewed as a trajectory for a 'clean' outcome (omitted from Figure \ref{diag:model}), of which a version $\mathbf{y_n}$ corrupted by Gaussian noise is observed.
Additive response functions can be seen as a continuous extension of scalar \emph{average treatment effect} (ATE) which is defined as the expected difference of outcomes before and after treatment.

\textbf{Response function:}
Response functions specify how treatments affect the outcome over time, and they should be specified to suit the application at hand, balancing flexibility, interpretability, etc. For example, if interpretability is not needed and the amount of data is large, non-parametric functions that learn the shape of the response are attractive. On the other hand, parametric functions are suitable when data are scarce, and they are often interpretable, which is valuable in itself but also helps specifying prior knowledge to improve accuracy. In the application of learning the impact of meals on blood glucose (Section \ref{ss:realdata}), we model the treatment response using a bell-shaped parametric function
\begin{equation}
    \pazocal{R}_{nm} \defeq f(\Delta_{nm}, h_{nm}, l_{nm}) \defeq h_{nm} \exp \left \{ \frac{-0.5(\Delta_{nm}-3l_{nm})^2}{l_{nm}^2} \right \},
\label{eq:response}
\end{equation}
where a lag vector $\Delta_{nm} = \mathbf{\tau}_n - t^{*}_{nm}$ represents the time since a specific treatment. The shape of this response is shown in Figure \ref{fig:trfunc} and it is determined by two parameters $h_{nm}$ and $l_{nm}$ with straightforward interpretations: $h_{nm}$ is the height of the response, and $l_{nm}$ is the length-scale which is proportional to the 'width' or 'duration' of the response. The main challenge in our application is scarceness and noisiness of data, with only 13 individuals and on average 10 meals per patient. We also tried a more flexible three-parameter response used in \cite{schulam2017reliable}, which allows a skewed response (see Figure \ref{fig:trfunc}), but this model suffered from convergence problems, for which reason we selected the simpler alternative.

In applications it is often of interest to measure how the response depends on treatment covariates, and therefore we allow these parameters to depend on the covariates:
\begin{align} 
\begin{split}
    h_{nm} &= (\beta^{h}_n)^T \textbf{x}^*_{nm}, \; \text{and} \\
    l_{nm} &= (\beta^{l}_n)^T \textbf{x}^*_{nm}, \; \text{for all } n,m. 
\end{split}\label{eq:height_width}
\end{align}
In Equation (\ref{eq:height_width}), the coefficient vectors $\beta^{h}_n,\beta^{l}_n \in \mathbb{R}^P$ represent the \textit{personalized impact} of each of the $P$ covariates on the height or width of the response for the $n$th individual. To share information across individuals, we introduce a Bayesian hierarchical prior, see \cite{gelman2013bayesian}, and assume that the personalized height and length-scale coefficients, $\beta^{h}_n$ and $\beta^{l}_n$, are drawn from common distributions:
\begin{equation*} 
    \beta^{h}_n \sim N_P(\tilde{\beta}_{h}, \Sigma_h) \quad \text{and} \quad \beta^{l}_n \sim N_P(\tilde{\beta}_{l}, \Sigma_l).
\end{equation*}
A hyperprior is further placed on the mean parameters of these distributions: 
\begin{equation*} 
    \tilde{\beta}_{h} \sim N_P(\bm{0}, \tilde{\Sigma}_h) \quad \text{and} \quad \tilde{\beta}_{l} \sim N_P(\bm{0}, \tilde{\Sigma}_l)
\end{equation*}
The hierarchical prior introduces shrinkage and facilitates estimation of the personalized coefficients with limited data. Further details are given in the Supplementary material.

\textbf{Counterfactual trend:} A counterfactual trend represents the outcome assuming no treatment has been taken. It has to be sufficiently flexible to handle any variation in the outcome that is not accounted for by the treatments. In this paper, we model the trend $\pazocal{T}_n(t)$ for individual $n$ using a Gaussian Process (GP) \cite{rasmussen2004gaussian}:
\begin{equation*}
    \pazocal{T}_n(t) \sim \pazocal{GP}(0, k(t,t'|\theta_{\pazocal{T}_n})),
\end{equation*}
where $\theta_{\pazocal{T}_n}$ are parameters associated with the kernel function $k(x,x'|\theta_{\pazocal{T}_n})$. GPs are non-parametric regression models with well-known closed-form formulas for posterior estimation, which they inherit from the Normal distribution by assuming all training and test data follow a joint Normal distribution. For example, if
\begin{equation*}
    \pazocal{S}_n = \textbf{y}_n - \sum_m\pazocal{R}_{nm}
\end{equation*}
is the residual of the outcome after subtracting the impact of the treatment responses, then
\begin{equation*} 
\begin{split} 
    \pazocal{T}_n(t)|\pazocal{S}_n &\sim N(\mu_*, \Sigma_*), \quad \text{where} \\
    \mu_* &= k(\mathbf{\tau_n},t)^T K(\mathbf{\tau_n},\mathbf{\tau_n})^{-1} \pazocal{S}_n, \quad \text{and}\\
    \Sigma_* &= k(t,t) - k(\mathbf{\tau_n},t)^T K(\mathbf{\tau_n},\mathbf{\tau_n})^{-1} k(\mathbf{\tau_n},t).
\end{split}
\end{equation*}
We refer the reader to \cite{rasmussen2004gaussian} for more details about GPs. As the kernel, we use the sum of Squared Exponential (SE) and constant kernels, where the former equips the GP with desired smoothness, and the latter enables meaningful extrapolation to regions where no input points have been observed. To speed up computation, we use a sparse GP \cite{rasmussen2004gaussian} instead of a full GP, which samples a small set of inducing points uniformly from $\mathbf{\tau_n}$ to achieve a low-rank approximation of $K(\mathbf{\tau_n},\mathbf{\tau_n})$ and its inverse. A detailed prior specification is provided in the Supplementary material.

\textbf{Measurement models:} Measurement models describe error in observations. With self-reported data both covariates and the timing of a treatment may be uncertain. To account for the uncertainty in treatment timing, we assume:
\begin{equation*}
    t_{nm} \sim N(t^{*}_{nm} + d_n, (\sigma_n^t)^2), \quad \text{for all } n,m.
\end{equation*}
In words, the observed time $t_{nm}$ is obtained from the true time $t^{*}_{nm}$ by shifting it with a bias term $d_n$, and adding Gaussian noise. The bias term $d_n$ represents reporting habits of different individuals. For example, in the blood glucose application in Section \ref{ss:realdata}, some individuals may systematically report their meal after eating, while others may do this before eating.

Different models are possible for treatment covariates, depending on the assumptions and data available \cite{MeasurementErrorBook}. Here we assume a simple perturbation on the \textit{amount} of treatment:
\begin{equation}
\begin{split}
    \mathbf{x}_{nm} &= \mathbf{x}^*_{nm} \delta_{nm}, \quad \text{where}\\ 
    \delta_{nm} &\sim LogNormal(0, \sigma_x^2), \quad \text{for all } n,m.
\end{split}\label{eq:error_in_cov}
\end{equation}
The coefficient $\delta_{nm}$ represents the error in the $m$th treatment of the $n$th individual. Intuition in the blood glucose application is that users are able to report correctly what they have eaten, but not how much. While the model (\ref{eq:error_in_cov}) captures our understanding of the type of mismeasurement expected in our data, more complicated models could also be justified, but they would require stronger additional assumptions to resolve the nonidentifiability of the EIV models. The model in (\ref{eq:error_in_cov}) is identifiable and can be trained with relatively little data, as we demonstate in Section \ref{Exp}.


Estimating $t^{*}_{nm}$ is straightforward as it only shifts the response, but does not change its shape. However, estimating $\mathbf{x}^*_{nm}$ is more complicated, and requires assuming that the counterfactual trend is sufficiently regularized. Otherwise the trend could easily compensate for the perturbation. We solve this by encouraging a large length-scale for the squared exponential kernel in the prior. Further details, e.g., prior distributions for $\mathbf{x}_{nm}^*$, $t_{nm}^*$, and $d_n$, are provided in the Supplementary material.

\subsection{A note on causality}\label{causality}
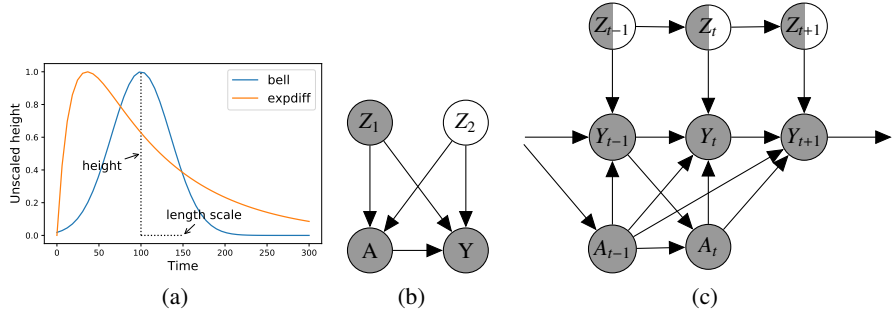
\begin{figure}[tb]
    \centering
    \subfloat[\mbox{}]
    	{
    		\makebox[.3\columnwidth]{
    	\centering
    	\includegraphics[width = .39\textwidth]{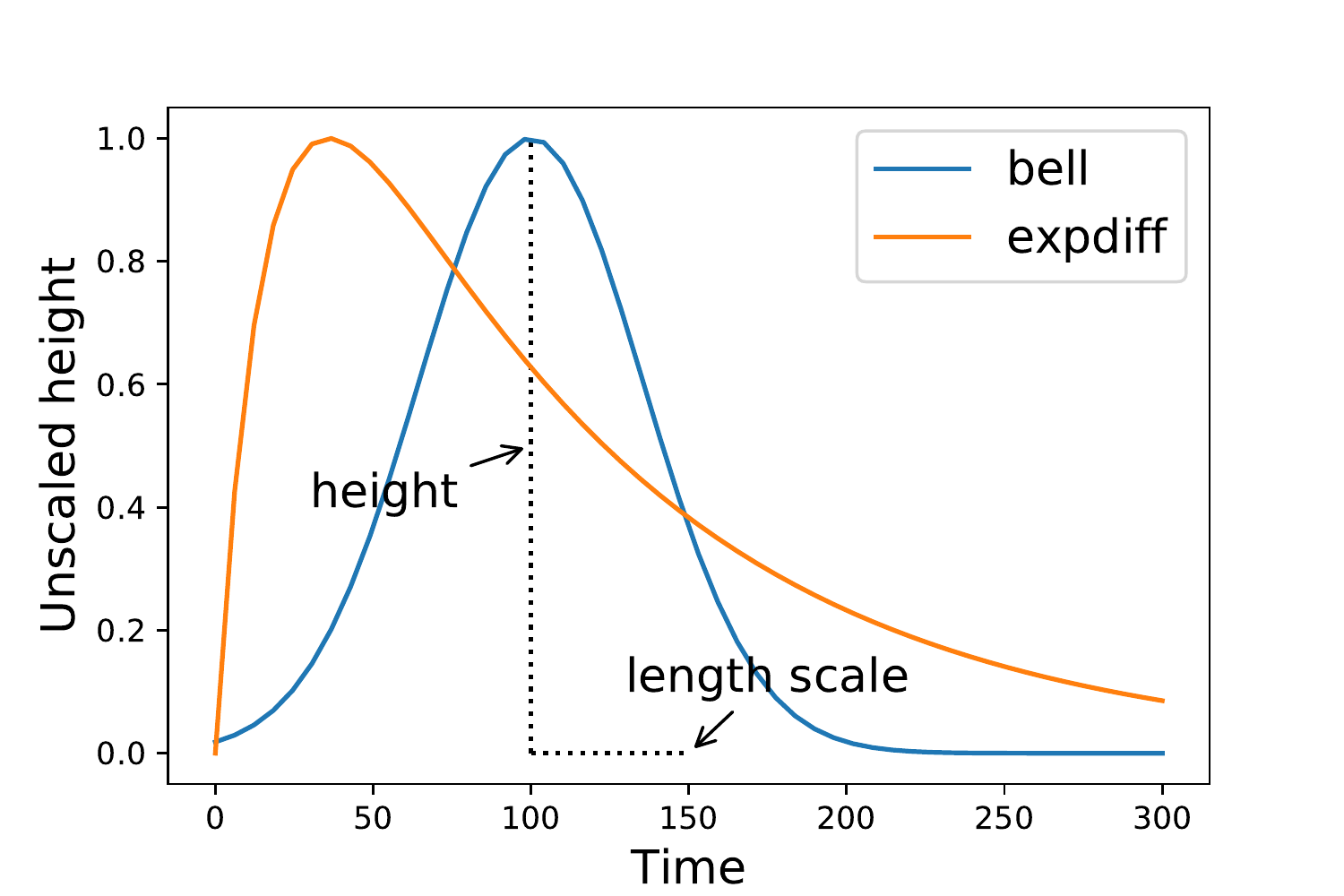}
    }
    	\label{fig:trfunc}} 
    	\hspace{2mm} 
	\subfloat[\mbox{}]
	{
		\makebox[.15\columnwidth]{
		\begin{tikzpicture} [scale=0.85, transform shape] 
		\node[obs,fill={darkgray}] (A) at (0, 1) {A}; 
		\node[obs,fill={darkgray}]  (Y) at (1.5, 1) {Y};
		\node[obs,fill={darkgray}]  (L) at (0, 3) {$Z_1$};
		\node[latent]  (Z) at (1.5, 3) {$Z_2$};
	    \edge {A} {Y} ; %
	    \edge {L} {Y} ; %
	    \edge {L} {A} ; %
	    \edge {Z} {Y} ; %
	    \edge {Z} {A} ; %
		\end{tikzpicture}
	}\label{DAG1}} 
	\subfloat[]{
		\makebox[.47\columnwidth]{ 
		\begin{tikzpicture} [scale=0.85, transform shape] 
		\node[obs,fill={darkgray}] (Y1) at (0, 2) {$Y_{t-1}$}; 
		\node[left=of Y1] (Q1) {};
		\node[obs,fill={darkgray}] (Y2) at (1.5, 2) {$Y_{t}$};
		\node[obs,fill={darkgray}] (Y3) at (3, 2) {$Y_{t+1}$}; 
		\node[right=of Y3] (Q2) {};
		\node[obs,below=of Y2,fill={darkgray}] (A1) {$A_{t}$};
		\node[obs,below=of Y1,fill={darkgray}] (A0) {$A_{t-1}$};
		\node (U1)[latent,above=of Y1,path picture={\fill[gray!80] (path picture bounding box.south) rectangle (path picture bounding box.north west);}] (U1) {$Z_{t-1}$};
		\node (U2)[latent,above=of Y2,path picture={\fill[gray!80] (path picture bounding box.south) rectangle (path picture bounding box.north west);}] {$Z_{t}$};
		\node (U3)[latent,above=of Y3,path picture={\fill[gray!80] (path picture bounding box.south) rectangle (path picture bounding box.north west);}] {$Z_{t+1}$};
		\edge {Q1} {Y1} ; %
		\edge {Y3} {Q2} ; %
	    \edge {Y1} {A1} ; %
	    \edge {Q1} {A0} ; %
	    \edge {A0} {Y1} ; %
	    \edge {A0} {Y2} ; %
	    \edge {A0} {Y3} ; %
	    \edge {A0} {A1} ; %
	    \edge {A1} {Y2} ; %
	    \edge {A1} {Y3} ; %
	    \edge {Y1} {Y2} ; %
	    \edge {Y2} {Y3} ; %
	    \edge {U1} {U2} ; %
	    \edge {U2} {U3} ; %
	    \edge {U1} {Y1} ; %
	    \edge {U2} {Y2} ; %
	    \edge {U3} {Y3} ; %
		\end{tikzpicture}
}\label{DAG2}} 	
\caption{a) Two Response functions. The blue one is used in this paper, while the orange one is used in \cite{schulam2017reliable}, (b) Graphical model for a cross-sectional case, showing action ($A$), response ($Y$), and observed and hidden confounders ($Z_1$ and $Z_2$), and (c) over-time response with a single treatment, where confounders $Z$ can be either observed or hidden.  
}
\end{figure}

We briefly review results related to estimation of causal effects from observational data on treatment-response trajectories \cite{pearl2009causality,hernan2018causalinference,schulam2017reliable}, to enable a user of our method to judge to what extent the effects found may or may not be interpreted causally. The causal effect of an action $A$ (e.g. a treatment) on $Y$ is defined as $P(Y=y|do(A=a))$, where the $do(\cdot)$ operator represents a manipulation of $A$ to value $a$. The key assumption is that there are \textit{no unmeasured confounders} (NUC), such as $Z_2$ in Figure \ref{DAG1}. Without $Z_2$, the causal effect of $A$ on $Y$ can be estimated from observational data using the adjustment formula:
\begin{equation*}
    P(Y=y|do(A=a))= \sum_{z_1} P(Y=y|A=a, Z_1=z_1)P(Z_1=z_1).
\end{equation*}
Time-varying treatments (Figure \ref{DAG2}) further face an issue of treatment-confounder feedback \cite{hernan2018causalinference}, which means that hidden confounders do not have to affect $A$ directly to create a spurious correlation between an action and future observations. A generalized adjustment formula, g-formula, can still be used to calculate $P(\bar{Y}_{\ge t}|do(A_{t-1},A_t))$, see \cite{daniel2013methods}.

A useful result, applicable with our model, is to use the model to estimate $P(\bar{Y}_{\ge t}|\bar{A}_{\le t}, \bar{Y}_{<t})$ from observational data. Then, assuming NUC, the following holds \cite{hernan2018causalinference}:
\begin{align}
P(\bar{Y}_{\ge t}|do(A_t), \bar{A}_{< t}, \bar{Y}_{<t}) = P(\bar{Y}_{\ge t}|\bar{A}_{\le t}, \bar{Y}_{<t}).\label{eq:causal_interpretation}
\end{align}
In words, conditionally on the history of treatments and the outcome (and relevant observed confounders not shown in the formula), the causal impact of the most recent treatment on future outcomes can be estimated from observational data. This \emph{short-term effect} can be used, e.g., to select between alternative treatments available at a certain point in time, when the relevant history of the individual is known.

In Section \ref{ss:realdata} we analyse the impact of diet on blood glucose. Based on domain knowledge, we know that diet is a prominent cause of changes in blood glucose. Furthermore, in our data we often see a rapid increase and decrease in blood glucose after a meal. Therefore, it seems plausible that meals affect blood glucose causally. However, in general, the causal assumptions can not be verified from observational data, and it is possible that some confounder affects both glucose and diet, but the effect of any such confounder is expected to be small compared to the impact of diet. Hence, while interpreting our results causally seems reasonable, we can not make assertions of this. More generally, with emergence of modern wearable self-monitoring devices, it will be possible to measure all relevant factors that could affect blood glucose much more comprehensively, and the NUC assumption is reasonable. Our model is straightforward to extend to such data.

\section{Experiments}
\label{Exp}
In this section, we first examine identifiability and accuracy of our model using simulated data, and then use it to analyze a real-world dataset comprising diet and continuous blood glucose measurements. 
Throughout, we compare four models, in an increasing order of complexity (later models include the previous as special cases):
\begin{itemize}
\item $\pazocal{M}_{ind}$ : Separate models for individuals.
\item$\pazocal{M}_{hier}$ : Model with the hierarchical prior for the responses to share information across individuals.
\item $\pazocal{M}_{hier+time}$ : Time uncertainty included.
\item $\pazocal{M}_{hier+time+cov}$ : Uncertainty in covariates included.
\end{itemize}

\subsection{Identifiability and accuracy of the models on simulated datasets} 

As a simple experiment, we first study the identifiability of the EIV model when there is measurement error in the covariates. We simulate artificial data using a toy model specified as the sum of a linear trend and the parametric treatment response from Equation (\ref{eq:response}). The dimension of treatment covariates is here set to 2, and each input is perturbed with an additive term drawn from $N(1,0.2^2)$. We analyze the data using the EIV model that assumes measurement error, and a model that disregards the noise in the covariates. Results and further details for this simple setup are presented in the Supplementary material, and they show that the EIV model recovers all true inputs and effect sizes with high accuracy, while the model that neglects the noise leads to biased coefficient estimates with wide confidence intervals.

To study the accuracy and identifiability of our method in a more realistic simulated setup, we first fit our model to the real-world data from Subsection \ref{ss:realdata}, and use the fitted model to simulate responses and trends for two individuals. We perturb half of the inputs according to Equation (\ref{eq:error_in_cov}) and let the model use the perturbed inputs and try to recover the true inputs and parameters. The performance of all models depends on the relative contributions of the trend and responses, and we scale up the response with a factor of 5, which facilitates a meaningful comparison.

\begin{figure*}[!htb]
	\centering
	\includegraphics[width = .99\textwidth]{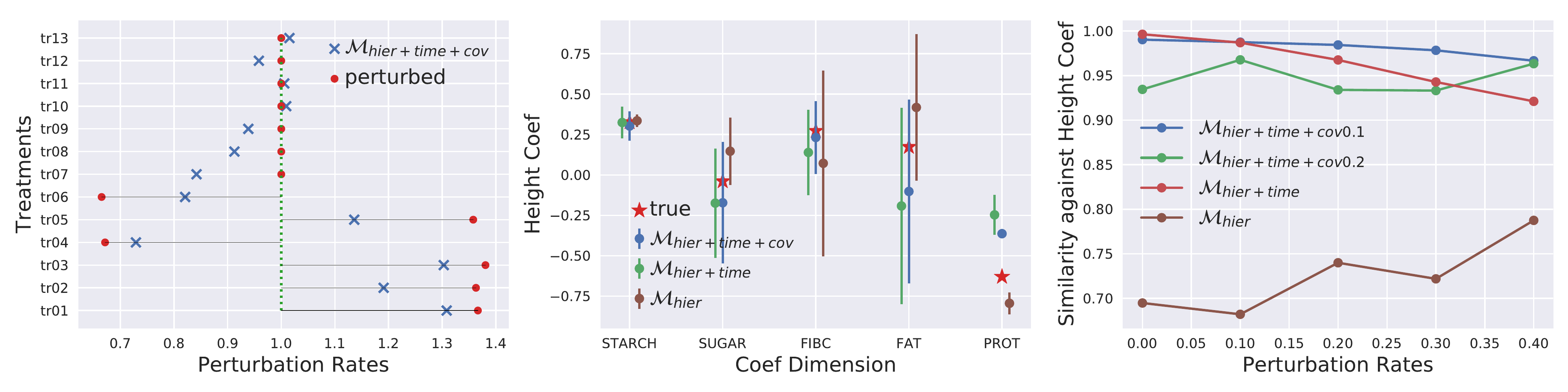}
	\caption{Simulation results. \textit{Left:} true and estimated perturbations for one individual (the other one shown in Supplement); \textit{Center:} true and estimated coefficients for the height of the response for 5 covariates; \textit{Right:} Cosine similarity of concatenated height coefficient vectors from all individuals against the true value with different levels of perturbation (higher value is better). Two different prior SDs, 0.1 and 0.2, were considered for model $\pazocal{M}_{hier+time+cov}$.}\label{fig:eiv_patients}
\end{figure*}

Results for one individual are shown in Figure \ref{fig:eiv_patients}, and for the other in the Supplementary material. We see that the direction of each non-zero perturbation is estimated correctly (left panel), and this is true also for the other individual (Supplementary). On the other hand, if there is no perturbation, the model may even then estimate non-zero perturbations, introducing additional noise. This reflects the trade-off between flexibility and overfitting, and highlights the importance of carefully validating the model to suit the amount and complexity of data. We also see that the regression coefficients are estimated accurately by the EIV model (center), and that the benefit from using EIV becomes more significant when the size of the perturbation increases (right).
However, a too loose EIV prior (large SD) may actually harm the performance by introducing additional noise, when the true perturbation is small.

\subsection{Experiments on real-world glucose data}\label{ss:realdata}
The data contain blood glucose measurements and dietary records. These anonymized data were provided by the Obesity Research Unit at the University of Helsinki, Finland, and they are available for $13$ non-diabetic individuals across three days. Diabetic individuals were excluded because their metabolism differs extensively from healthy individuals, and detailed modeling of that is beyond the scope of this work. The real-valued blood glucose measurements were collected by a portable continuous glucose monitoring system approximately every fifteen minutes. The dietary records consist of user-reported contents and times of all meals eaten during the 3-day study period. Each meal has been processed into amounts of five nutrients: starch, sugar, fiber, fat, and protein. The goal of the analysis is to learn how these nutrients influence blood glucose. Both the exact amounts of food eaten and the exact meal times are uncertain, as they are based on values estimated and reported by users, which motivates the use of EIV models for these data. A visualization of the data (and results) for one individual is shown in Figure \ref{fig:prediction}, and for all other individuals in the Supplement. Some markers may be missing due to device errors or when a user has removed the device.

\begin{figure*}[!htb]
	\centering
	\includegraphics[width = .99\textwidth]{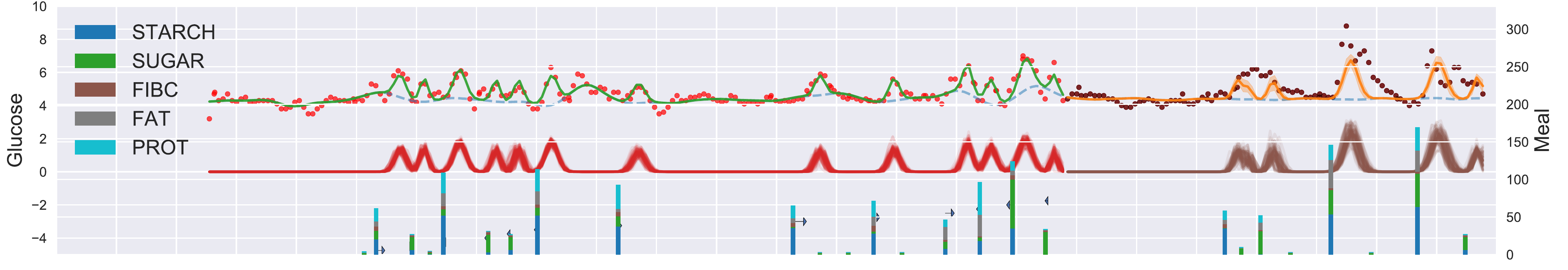}
	\caption{Visualization of the 3-day time series for one patient. 
	Red and brown dots represent glucose markers in the training and test sets, respectively. Meals are indicated by vertical bars, colored according to amounts of different nutrients in the meal. The green curve is the final fitted trajectory, and it is a combination of the dashed blue line, a counterfactual trend, and the mean of red lines, which are posterior samples of treatment responses. Horizontal arrows associated with the meals show the estimated difference between true and observed meal times.
	}\label{fig:prediction}
	\captionsetup{justification=centering}
\end{figure*}

\begin{figure*}[!htb]
    \centering
    \captionsetup[subfloat]{farskip=0.5pt,captionskip=0.5pt}
    \subfloat{
    	\makebox[.99\textwidth]{
        \includegraphics[width = .99\textwidth]{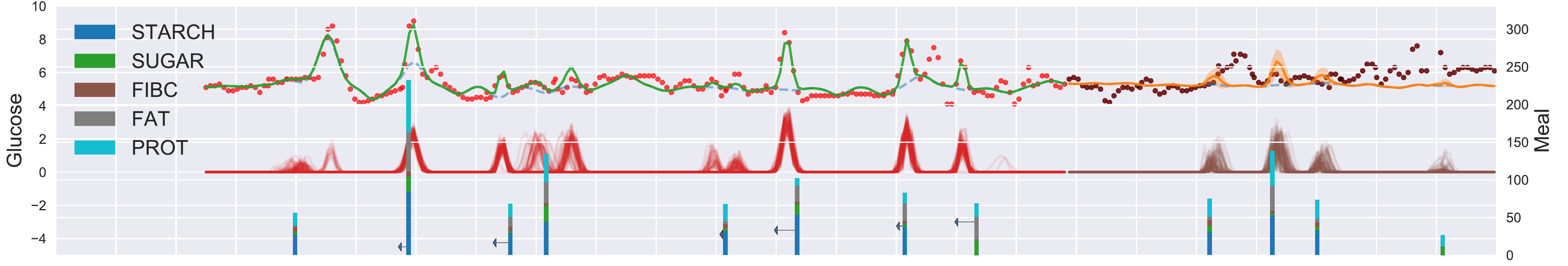}
    }}
    \quad
    \subfloat{
    	\makebox[.99\textwidth]{
        \includegraphics[width = .99\textwidth]{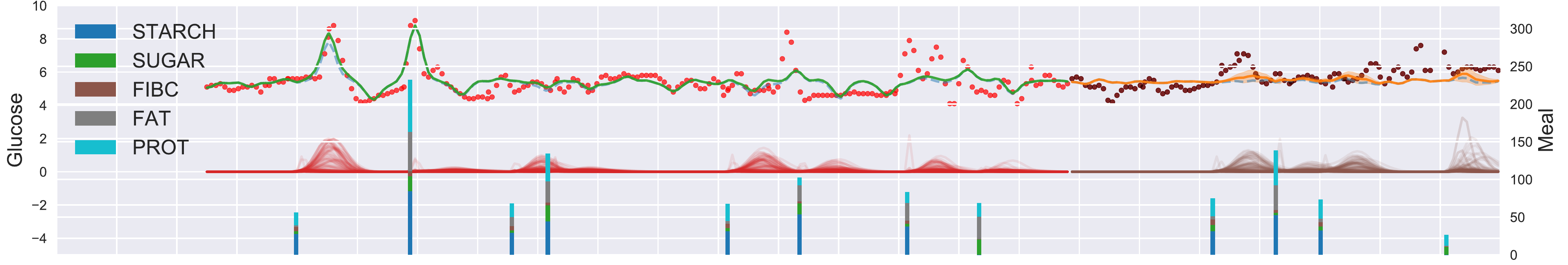}
    }}
    \caption{Demonstration of time uncertainty modeling for one individual. \textit{Upper:} Results using $\pazocal{M}_{hier+time}$, where arrows indicate the estimated difference between the true and observed meal times; \textit{Bottom:} Results using $\pazocal{M}_{hier}$.}\label{fig:time}
\end{figure*}

\textbf{Metrics:}
The models are trained using data from the first two days, and the third day is used for testing. The performance of treatment-response estimation is quantified using five metrics $\text{M}_i, i\in\{1,\ldots,5\}$. $\text{M}_1$ is the proportion of variance explained by the trend:
\begin{equation*}
    \text{M}_1 = \frac{1}{N}\sum_n \frac{\text{Var}(\pazocal{T}_n)} {\text{Var}(\mathbf{y}_n)}.
\end{equation*}
$\text{M}_2$ indicates how much more is explained when also the treatment responses are included:
\begin{equation*}
    \text{M}_2 = \frac{1}{N}\sum_n \frac{\text{Var}(\pazocal{T}_n + \sum_m \pazocal{R}_{nm})} {\text{Var}(\mathbf{y}_{n})} - \text{M}_1.
\end{equation*}
In detail, a large $\text{M}_1$ means that the outcome is mostly explained by the trend, and a small $\text{M}_2$ represents an inactive treatment response. These metrics are computed in regions of non-zero treatment response.
Metrics $\text{M}_3$ and $\text{M}_4$ are simply the mean squared errors in the training and test data. They are calculated for all individuals for whom $\text{M}_2$ indicates that the response has been properly learned. Thus one patient, shown in Figure \ref{fig:time}, with $\text{M}_2\approx 0.05$ for the baseline model $\pazocal{M}_{hier}$ is excluded from MSE calculations (other patients have $\text{M}_2 > 0.3$).

Because $\text{M}_4$ measures pointwise error, it may be misleadingly low when the response shape is correct if its location is inaccurate. Metric $\text{M}_5$ is insensitive to the inaccuracy of location, and it measures the absolute error in variance between predicted response and outcome:
\begin{equation*}
    \text{M}_5 = \frac{1}{N}\sum_n |\text{Var} \Big( \sum_m \pazocal{R}_{nm} \Big) - \text{Var}(\mathbf{y}_{n})| 
\end{equation*}
Because our interest is in estimation of the treatment response, and not the trend, we calculate $\text{M}_4$ and $\text{M}_5$ in windows from one hour before to three hours after each meal.

We use the Mann–Whitney U-test \cite{mann1947test} to test if other models are better than $\pazocal{M}_{hier}$ in terms of test error $\text{M}_4$. The reason for using $\pazocal{M}_{hier}$ as the baseline is the main argument of this article that EIV modeling is beneficial when estimating treatment-response trajectories, and $\pazocal{M}_{hier}$ is otherwise the same as $\pazocal{M}_{hier+time}$ and $\pazocal{M}_{hier+time+cov}$ except that it the does not include the EIV components. We also compare the models using an information criterion for predictive accuracy. The state-of-the-art criterion is leave-one-out cross-validation (LOO) \cite{vehtari2017practical}, which is used here.

\begin{table*}[h]
\centering
\footnotesize
\npdecimalsign{.}
\nprounddigits{3}
    \begin{tabular}{l|HHn{1}{3}|n{1}{3}|n{1}{3}|n{1}{3}|n{1}{3}|c|c|c|c|}
         &{old $\text{M}_1$} &{old $\text{M}_2$} 
        & \multicolumn{1}{p{1cm}|}{\centering $\text{M}_1$\\ PVE Trend}
        & \multicolumn{1}{p{1cm}|}{\centering $\text{M}_2$\\ PVE Resp}
        & \multicolumn{1}{p{1cm}|}{\centering $\text{M}_3$\\ MSE Train}
        & \multicolumn{1}{p{1cm}|}{\centering $\text{M}_4$\\ MSE Test}
        & \multicolumn{1}{p{1cm}|}{\centering $\text{M}_5$\\ $\Delta Var$ Test}
        & \multicolumn{1}{p{1cm}|}{\centering p-value\\ U-test}
        &LOO &pLOO 
        & \multicolumn{1}{p{1cm}|}{\centering SE\\LOO}\\
        \hline
        $\pazocal{M}_{ind}$ &2.429494 &0.516090 &0.360768 &0.342478 &0.149207 &1.695298 &0.926833 &1.00 &3549.64 &246.64 &318.8\\
        $\pazocal{M}_{hier}$ &2.618327 &0.473402 &0.359423 &0.339017 &0.158968 &0.751853 &0.390517 &- &3587.87 &214.62 &317.28\\
        $\pazocal{M}_{hier+time}$ &2.589473 &0.615982 &0.349797 &0.424396 &0.097750 &{\npboldmath}0.738217 & 0.376680 &\bf{3.24e-4} &\bf{2869.91} &342.24 &265.09\\
        $\pazocal{M}_{hier+time+cov}$ &2.604928 &0.615494 &{\npboldmath}0.344312 &{\npboldmath}0.427812 &{\npboldmath}0.097645 &0.743397 &{\npboldmath}0.365714 &4.66e-3 &2994.98 &465.47 &333.7\\  \hline 
    \end{tabular}
    \caption{Comparison of models using the real-world glucose data. The metrics $M_1$ through $M_5$ are defined in text, where PVE means \emph{Proportion of Variance Explained}. p-value tests if other models are better than $\pazocal{M}_{hier}$ in terms of $M_4$. LOO stands for leave-one-out cross-validation, pLOO is the estimated effective number of parameters, and SE-LOO records the standard error in the LOO computations.}\label{metrics}
\end{table*}

\textbf{Results:}
Result are shown in Table \ref{metrics}. We see that all models outperform the non-hierarchical baseline $\pazocal{M}_{ind}$ by a large margin. Furthermore, taking treatment time inaccuracy into account in $\pazocal{M}_{hier+time}$ improves significantly over the non-EIV model $\pazocal{M}_{hier}$. In fact, estimation of the response fails completely for some individuals without time EIV; the results with and without time uncertainty modeling for one such case are shown in Figure \ref{fig:time}. On the other hand, taking uncertainty in covariates into account does not notably improve accuracy, which is likely caused by a combination of increased flexibility and limited amount of data. Overall, models with EIV component outperform the model without EIV in all metrics.

Interpretability of personalized treatment response is also of great interest; for instance, understanding how an individual's glucose level changes if she eats one more unit of sugar. The overall goal of glucose monitoring is to keep the glucose level in a given range, and both the amount of excess and the duration of the hyperglycemic state are clinically important. Hence, a sensible parameter to consider is the impact of different nutrients on the \textit{area} of the response curve. Though this is not a parameter of our model, it is straightforward to derive the personalized increase in response area due to one unit increase of a specific nutrient $\Delta A_{np}$ ($n \in 1,\ldots,N$, $p \in \{1...P\}$), using coefficients for height and width, which are modeled explicitly (see Supplement).

\begin{figure}[h!]
    \centering
    \captionsetup[subfloat]{farskip=1pt,captionskip=0.5pt}
    \subfloat[]{
        \makebox[.25\textwidth]{
        \footnotesize
        \npdecimalsign{.}
        \nprounddigits{2}
        \begin{tabular}[b]{l|n{2}{2}n{2}{2}} \hline
& \text{Mean} & \text{Std} \\ \hline
STARCH&	9.573785&	4.288859\\
SUGAR&	6.363213&	4.966396\\
FIBC&	2.825785&	4.949791\\
FAT&	5.286884&	5.494602\\
PROT&	-10.129712&	3.898830\\\hline
        \end{tabular}
        \hspace{5mm}
    }\label{tbl:coef}}
    \quad
    \subfloat[]{
        \makebox[.3\textwidth]{
        \includegraphics[width = .3\textwidth]{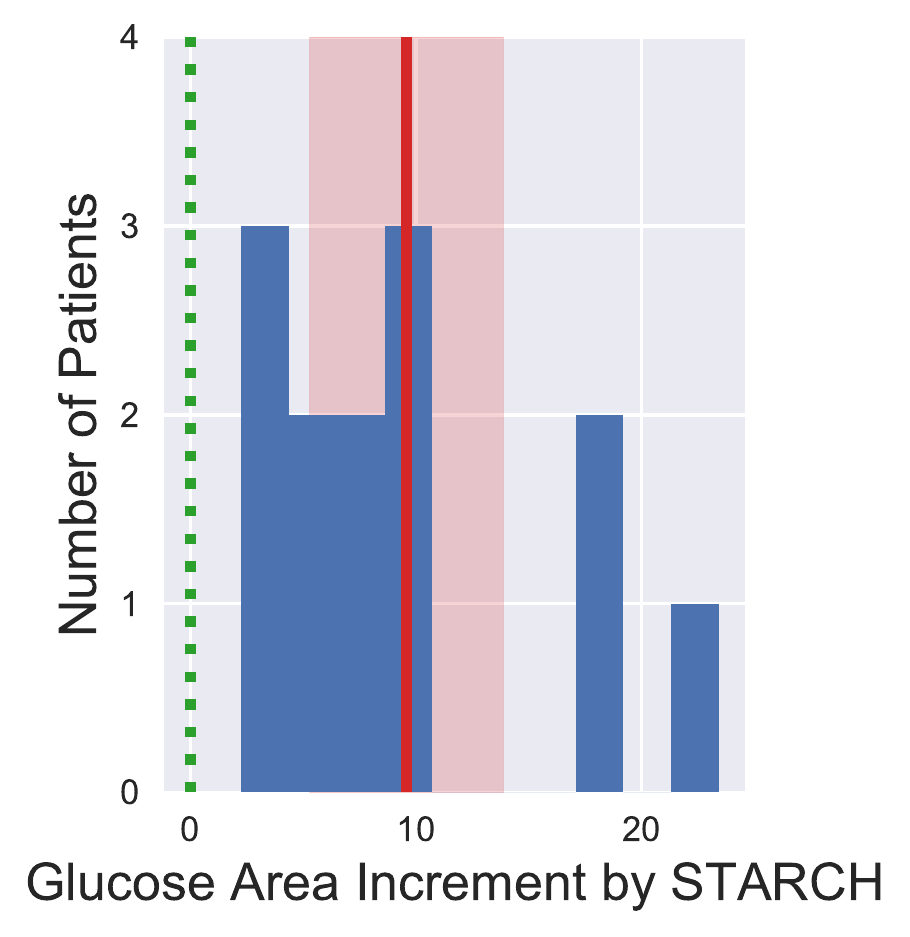}
    }\label{fig:coef}}
    \quad
    \subfloat[]{
    	\makebox[.35\textwidth]{
        \includegraphics[width = .4\textwidth]{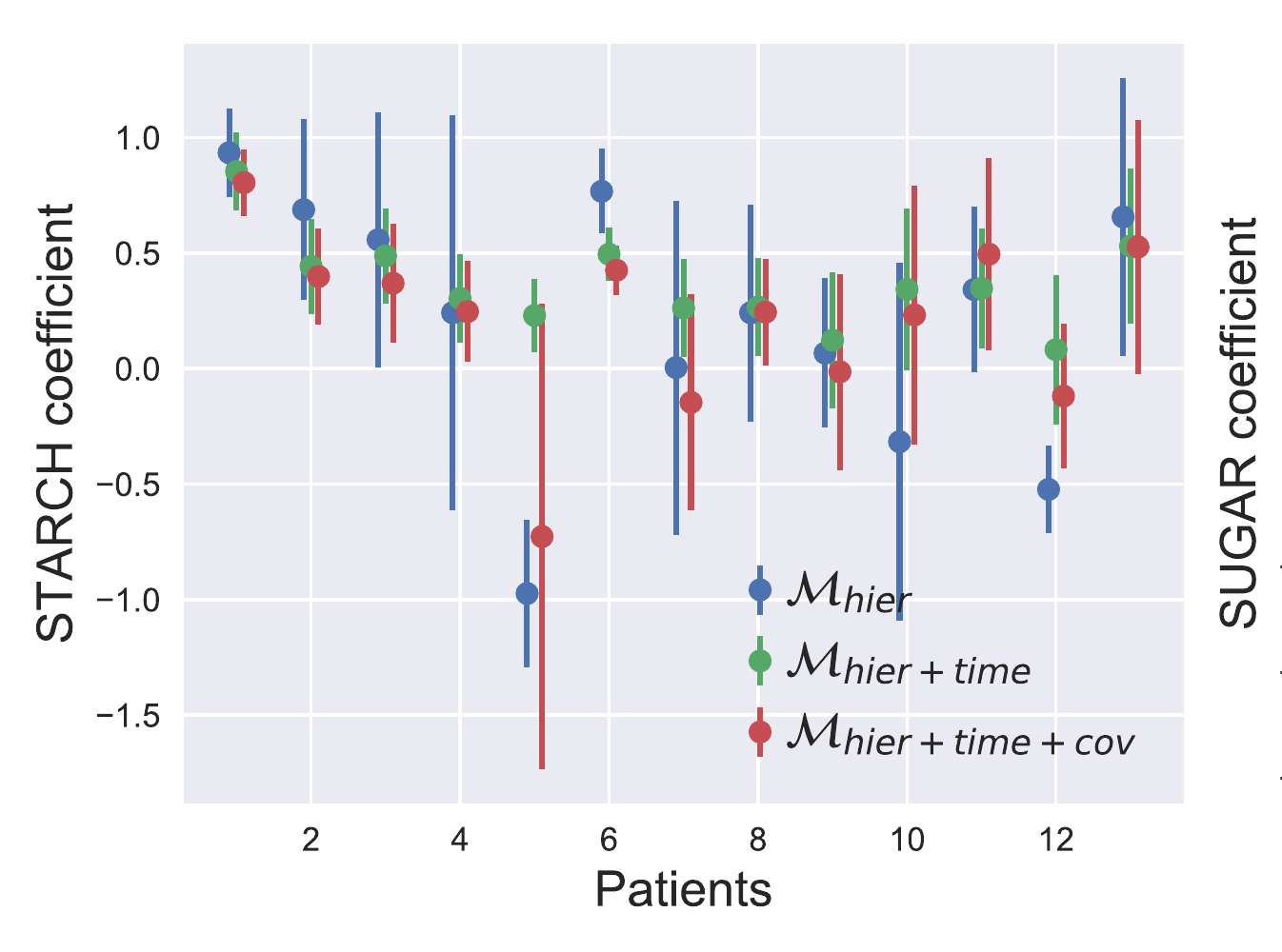}
    }\label{fig:coef_errorbar}}
    \caption{
    \textit{a}). Average impact on response area $\Delta A_{np}$ by different nutrients;
    \textit{b}) Histogram of personalized starch coefficients and their mean (+/- one SD) (red); \textit{c}) Posterior uncertainty in the personalized starch coefficients.
    }
\end{figure}

Overall, starch and sugar have the strongest positive impact on  glucose (Figure \ref{tbl:coef}), consistent with the understanding that carbohydrates increase blood glucose \cite{Wolever1995}. Protein, on the other hand, has a negative impact, which has been observed before and might represent a complex short-term interaction between nutrients \cite{Karamanlis2007}. An advantage of our model is that we get \textit{personalized} coefficients for each individual, as shown for starch in Figure \ref{fig:coef}, and for the other nutrients in the Supplement. Finally, posterior uncertainty of personalized starch coefficients is shown in Figure \ref{fig:coef_errorbar}. Importantly, models with EIV have much narrower confidence intervals, meaning that they are estimated more accurately, thanks to increased flexibility that allows fitting the complex data.

\section{Conclusion}
We presented a hierarchical model with EIV components to estimate personalized treatment-response trajectories when the covariates and timing of a treatment are imprecise. Our model demonstrates superior performance in both simulated and real-world data on various metrics, and allows extracting interpretable and meaningful estimates of the personalized impacts of treatment covariates, valuable in applications. Future directions include extensions and identifiability of EIV modelling, and extending the model to include interactions between covariates and other unmeasured confounders, such as physical activity, for causal completeness.

%
%
%
\bibliographystyle{splncs04}
\bibliography{main}
\end{document}